\documentclass[lettersize,journal]{IEEEtran}
\usepackage{amsfonts}
\usepackage{array}
\usepackage[caption=false,font=normalsize,labelfont=sf,textfont=sf]{subfig}
\usepackage{textcomp}
\usepackage{stfloats}
\usepackage{url}
\usepackage{verbatim}
\usepackage{graphicx}
\usepackage{cite}
\usepackage{times}
\usepackage{colortbl}
\usepackage{epsfig}
\usepackage{graphicx}
\usepackage{amsmath}
\usepackage{amssymb}
\usepackage{makecell} 

\usepackage{enumitem} 
\usepackage[linesnumbered,ruled,vlined,algo2e]{algorithm2e}
\usepackage{algorithm, algorithmic}
\usepackage{multirow}
\usepackage{threeparttable}
\usepackage{stfloats}
\usepackage{booktabs}
\usepackage{arydshln}
\usepackage{bm}
\usepackage{makecell}
\usepackage{pifont}
\usepackage{lipsum}
\usepackage{ulem}
\usepackage[table,dvipsnames]{xcolor} 
\usepackage[pagebackref,breaklinks,colorlinks]{hyperref}

\usepackage[capitalize]{cleveref}
\crefname{section}{Sec.}{Secs.}
\Crefname{section}{Section}{Sections}
\Crefname{table}{Table}{Tables}
\crefname{table}{Tab.}{Tabs.}

\colorlet{colorFst}{Green!25}       
 \colorlet{colorSnd}{SpringGreen!65} 
\colorlet{colorTrd}{Yellow!30}      
\colorlet{colorLow}{darkgray!30}    
\definecolor{darkpastelgreen}{rgb}{0.01, 0.75, 0.24}
\definecolor{darkgreen}{rgb}{0.00, 0.8, 0.2}
\definecolor{darkyellow}{rgb}{0.96, 0.75, 0.00}
\definecolor{badcolor}{rgb}{0.82,0.25,0.12}
\definecolor{ours}{rgb}{0.92, 0.92, 0.92}
\newcommand{\fs}{\cellcolor{colorFst}\bf}   
\newcommand{\nd}{\cellcolor{colorSnd}}      
\newcommand{\rd}{\cellcolor{colorTrd}}      
\newcommand{\imp}{\color{Green}}

\newcommand{\ours}{\cellcolor{ours}}

\def\eg{\emph{e.g.}\xspace} 
\def\ie{\emph{i.e.}\xspace}

\let\titleold\title
\renewcommand{\title}[1]{\titleold{#1}\newcommand{\thetitle}{#1}}

   \definecolor{baselinecolor}{gray}{.9}

\newcommand{\fadedtext}[1]{\textcolor{gray}{#1}}

\newcolumntype{x}[1]{>{\centering\arraybackslash}p{#1pt}}
\newcolumntype{y}[1]{>{\raggedright\arraybackslash}p{#1pt}}
\newcolumntype{z}[1]{>{\raggedleft\arraybackslash}p{#1pt}}
\newlength\savewidth

\renewcommand{\paragraph}[1]{\vspace{1.25mm}\noindent\textbf{#1}}

\begin{document}

\title{Trajectory Entropy: Modeling Game State Stability from Multimodality Trajectory Prediction}


\author{Yesheng Zhang*,~Wenjian Sun*,~Yuheng Chen,~Qingwei Liu,~Qi Lin,~Rui Zhang,~Xu Zhao
\thanks{* indicates equal contribution.}
\thanks{Yesheng Zhang, Wenjian Sun,~Yuheng Chen and Xu Zhao are with Department of Automation, Shanghai Jiao Tong University. (e-mail: {\tt\small \{preacher, sun.wen.jian,chenyuheng,zhaoxu\}@sjtu.edu.cn})
}
\thanks{Qingwei Liu, Qi Lin and Rui Zhang are with SAIC-ZONE~Technology~Co.~Ltd. (e-mail: {\tt\small \{zhangrui07, linqi01,liuqingwei01\}@saicmotor.com})
}
\thanks{Corresponding author: Xu Zhao.}}
%

\markboth{Journal of \LaTeX\ Class Files,~Vol.~14, No.~8, August~2021}%
{Shell \MakeLowercase{\textit{et al.}}: A Sample Article Using IEEEtran.cls for IEEE Journals}

\IEEEpubid{0000--0000/00\$00.00~\copyright~2021 IEEE}

\maketitle

\begin{abstract}
Complex interactions among agents present a significant challenge for autonomous driving in real-world scenarios.
Recently, a promising approach has emerged, which formulates the interactions of agents as a \textit{level-k} game framework.
It effectively decouples agent policies by hierarchical game levels. 
However, this framework ignores both the varying driving complexities among agents and the dynamic changes in agent states across game levels, instead treating them uniformly. 
Consequently, redundant and error-prone computations are introduced into this framework. 
To tackle the issue, this paper proposes a metric, termed as \textit{Trajectory Entropy}, to reveal the game status of agents within the level-k game framework.
The key insight stems from recognizing the inherit relationship between agent policy uncertainty and the associated driving complexity.
Specifically, Trajectory Entropy extracts statistical signals representing uncertainty from the multimodality trajectory prediction results of agents in the game.
Then, the signal-to-noise ratio of this signal is utilized to quantify the game status of agents. 
Based on the proposed Trajectory Entropy, we refine the current level-k game framework through a simple gating mechanism, significantly improving overall accuracy while reducing computational costs. 
Our method is evaluated on the Waymo and nuPlan datasets, in terms of trajectory prediction, open-loop and closed-loop planning tasks. The results demonstrate the \textit{state-of-the-art} performance of our method, with precision improved by up to $19.89\%$ for prediction and up to $16.48\%$ for planning.
\end{abstract}

\begin{IEEEkeywords}
Trajectory prediction, Autonomous driving, Agent Planning
\end{IEEEkeywords}

\section{Introduction}\label{sec:intro}

\IEEEPARstart{J}{oint} trajectory prediction and ego vehicle planning has been demonstrated as a promising approach to achieve intelligent Autonomous Driving (AD)~\cite{joint0,joint1,gameformer,dipp,dtpp}. The approach takes an end-to-end fashion, enhancing the interaction between agents, which is a desirable way for autonomous driving~\cite{uniad}.

However, precise trajectory prediction and effective planning remain highly challenging in complex driving environments. This primarily arises from the tightly coupled nature of the AD task, that is, the trajectory of one vehicle is influenced by other vehicles, and vice versa. 
To address the tight-coupling issue, GameFormer~\cite{gameformer} draws inspiration from level-k game theory~\cite{game} and treats ego planning on par with the trajectory prediction. It incorporates all interested agents into a hierarchical game framework with k levels.
Within each level, the Multimodality Trajectory Prediction (MTP) is performed for every agent.
At its core, the prediction of each agent at every level only receives the game results of other agents from the preceding level as input, in line with the principles of level-k game theory. 
Consequently, this framework effectively disentangles agent interactions into different game levels,  facilitating intelligent autonomous driving.

\begin{figure}[!t]
\centering
\includegraphics[width=\linewidth]{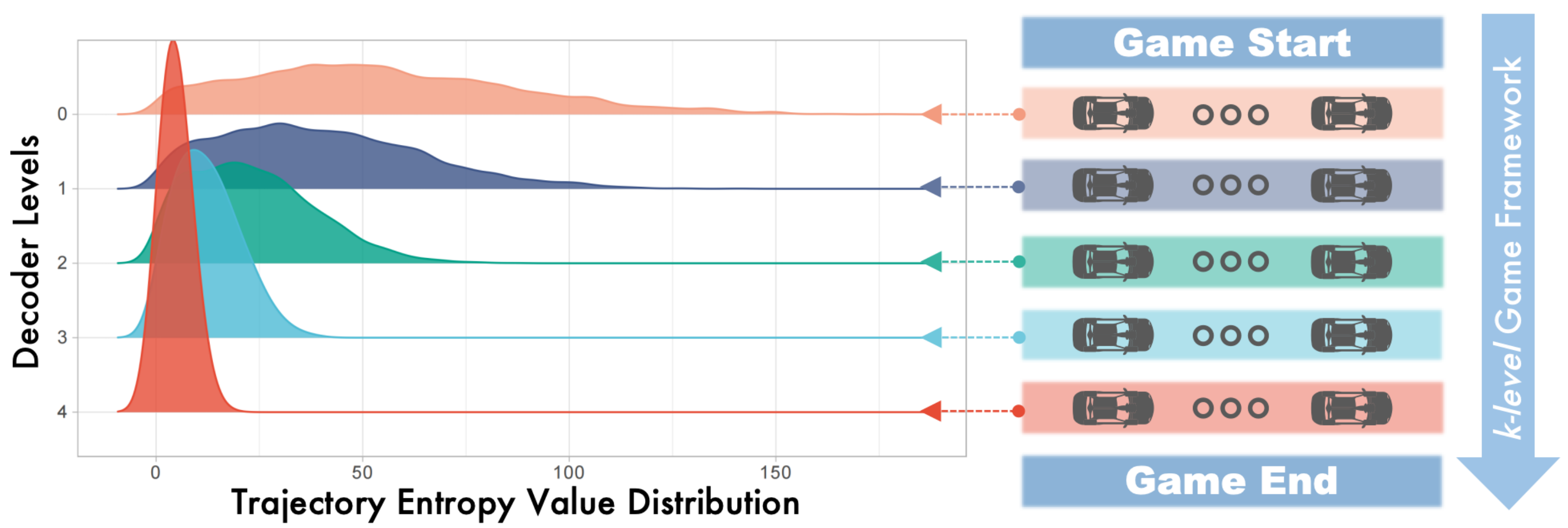}
\caption{\textbf{The visualization of entropy decrease in the level-k game, 
utilizing the proposed Trajectory Entropy.} We trained a GameFormer model on the Waymo dataset and employed it to predict trajectories in some random test scenes. Then, we analyzed the Trajectory Entropy (TE) values of GameFormer decoders at the five game levels. The outcomes reveal a decrease in TE as the game level ascends, suggesting improved overall stability with deeper game levels, in line with theoretical predictions of level-k game. These results underscore the efficacy of the proposed TE metric in capturing the game states of the agents.}
\vspace{-1.6em}
\label{fig:te-ana}
\end{figure}

\begin{figure*}[!t]
\centering
\includegraphics[width=\linewidth]{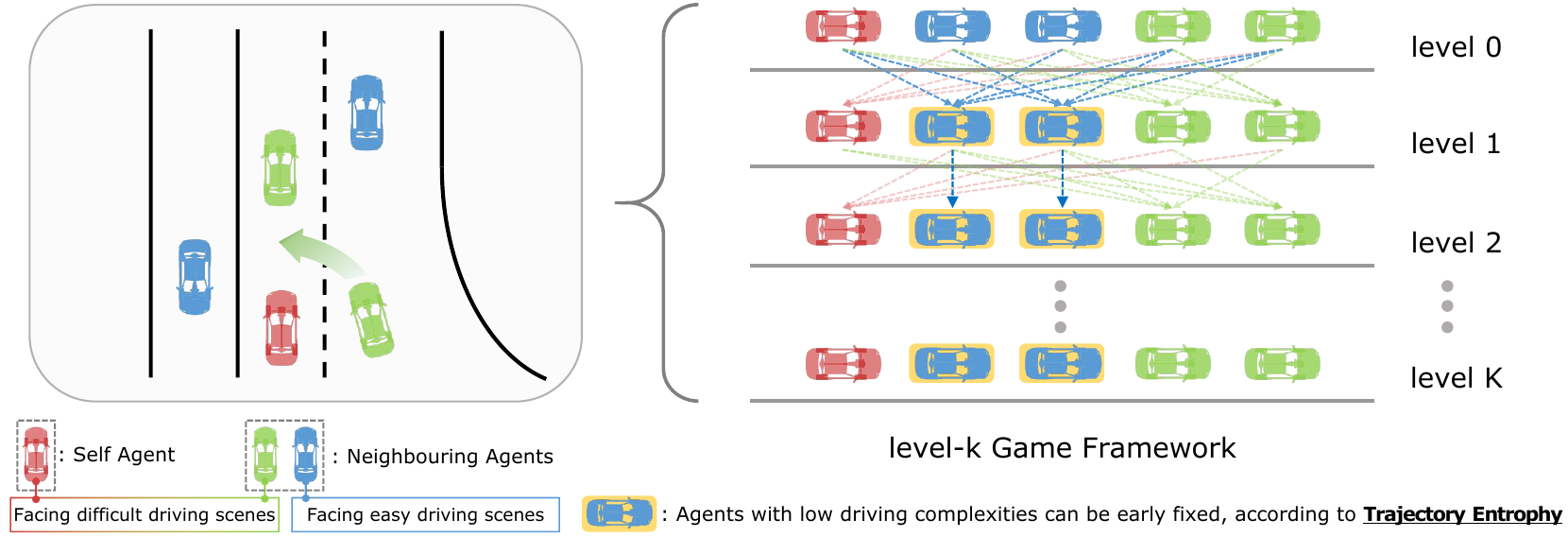}
\caption{\textbf{The motivation of our Trajectory Entropy.} In driving scenes, agents can face different driving complexities, as illustrated by the diverse vehicles in the diagram. Consequently, in the hierarchical game framework, these agents require tailored depths of reasoning. To this end, we propose \textit{Trajectory Entropy} to measure the driving difficulty of agents, enabling the allocation of appropriate game depths to each agent.}
\vspace{-1.6em}
\label{fig:kg}
\end{figure*}

Nevertheless, in the hierarchical game framework, treating all agents equally like GameFormer could introduce computational redundancy. 
In a given driving scene, agents may face varying levels of decision-making complexity due to their spatial positions and surrounding environments, such as agents in \cref{fig:kg}. 
This leads to diverse complexities of different agents within the game framework.
For example, some agents with low driving difficulties may only require a few levels of games to determine their future trajectories. However, high-difficulty agents may need more levels of games for deeper reasoning about their policy.
Therefore, including agents with already stable game states in unnecessary reasoning levels not only increases computational overhead but also introduces noise into the overall framework, thereby negatively impacting the final accuracy. 
Hence, accessing the game state of agents and allocating different game levels to different agents are needed for this level-k game framework.
\IEEEpubidadjcol

To address this, this work introduces a measurement named \textit{Trajectory Entropy}, which can effectively represent agent state stability in level-k game (cf. \cref{fig:te-ana}). The key insight lies in that the MTP results in each game round inherently reflect the corresponding game state. Particularly, MTP aims to provide multiple reasonable prediction outcomes to cover all feasible trajectories~\cite{MTP_survey}. In the context of autonomous driving, the distribution of feasible trajectories for each agent potentially reflects the level of driving complexity they encounter. 
For instance, if few agent interactions exist, the distribution of feasible trajectories for agents is naturally concentrated near the road's centerline, according to the traffic rules, indicating low driving difficulty (cf. blue cars in \cref{fig:kg}). 
On the other hand, in complex scenes with more agent interactions, feasible trajectory distribution of a agent is more dispersed, requiring multiple game rounds (\ie, negotiations) with other agents (cf. red and green cars in \cref{fig:kg}). 
Thus, the MTP results, which model the feasible trajectory distribution, are able to offer clues of the agent states in the level-k game.
Based on this, \textit{Trajectory Entropy} evaluates the game states of agents by measuring the dispersion of MTP results, enabling the elimination of redundant and error-prone computation in the level-k game framework.
This ultimately enhances the performance of trajectory prediction and planning.

Specifically, inspired by random signal processing theory~\cite{prs}, \textit{Trajectory Entropy} considers the distances among trajectories from different modalities as signals reflecting the degree of dispersion.
It also links trajectory confidence to the random noise accompanying this signal. 
Then, the signal-to-noise ratio (SNR) of this random signal could reflect the degree of dispersion in MTP, thereby revealing the game states of agents in the level-k game. 
As the magnitude of the signals is proportional to the confusion level of predicted trajectories, drawing on the concept of entropy from information theory, this metric is named as \textit{Trajectory Entropy}.

Then, we integrate the \textit{Trajectory Entropy} into the level-k game framework of GameFormer, aiming to reduce computational costs and enhance accuracy at the same time. 
In practice, \textit{Trajectory Entropy}s are computed from the MTP results for agents in the first level of the game. 
Subsequently, agents with \textit{Trajectory Entropy}s below a certain threshold are considered to have stable game states.
Then, a simple gate is employed to fix them in subsequent games, avoiding redundant computations on these agents and reducing uncertainty in subsequent games as well (\cref{fig:kg} right).
For unstable agents, their \textit{Trajectory Entropy} values are repeatedly evaluated at each game level to determine whether to fix them, until the final level.
With this gate module, the performance of GameFormer can be effectively enhanced.

Currently, GameFormer is the only algorithm that adopts the \textit{level-k} game framework for joint trajectory prediction and decision-making.
However, driving behavior inherently involves policy negotiation among agents, necessitating a game-theoretic framework.
Therefore, the proposed \textit{Trajectory Entropy}, as a general indicator of game state stability, is expected to be applicable to a broader range of game-theoretic frameworks in the future.

We summarize our main contributions as follows:
\begin{enumerate}
	\item Observing the varying levels of driving complexities faced by different traffic participants, we propose treating agents differently within the level-k game framework of GameFormer. By adjusting the reasoning depth of agents based on their game states, we aim to reduce computational redundancy and enhance the accuracy of joint trajectory prediction and planning.

	\item We introduce \textit{Trajectory Entropy} as a bridge from MTP to the game states of agents. Drawing inspiration from signal processing theory, we extract signals reflecting uncertainty from MTP, to serve as a clue for revealing the driving difficulties of agents. \textit{Trajectory Entropy} can be used to assess the game states of agents, thereby enabling the improvement of the level-k game framework.

	\item Through a simple Trajectory Entropy Gate mechanism, we effectively improve the level-k game framework of GameFormer. This approach has demonstrated \textit{state-of-the-art} results in trajectory prediction, open-loop and closed-loop planning on the WOMD~\cite{waymo} and NuPlan~\cite{nuplan} benchmarks.
\end{enumerate}

\section{Related Work}

\subsection{Multimodal Trajectory Prediction}
Trajectory prediction is a fundamental task in modern autonomous driving systems. Its objective is to anticipate the future paths of agents in the scene, providing information for planning model to ensure safe and intelligent driving.
Initially, the goal of trajectory prediction was to forecast a single trajectory for the agent, referred to as discriminative trajectory prediction~\cite{MTP_survey}. However, due to the inherent uncertainty of agent behavior, multiple feasible trajectories exist from start to destination. In such cases, discriminative trajectory prediction models struggle to differentiate between these feasible solutions.
To address this uncertainty, multimodal trajectory prediction tasks have been introduced~\cite{socialgan}, requiring models to output multiple prediction trajectories to cover all feasible solutions. Owning to its interpretability, multimodal output has now become a default setting in trajectory prediction for autonomous driving tasks.

In order to enable models to produce multiple feasible trajectory predictions in complex environments, deep learning models are adopted in most current approaches.
Initially, Long Short-Term Memory networks (LSTM)~\cite{sociallstm} and Convolutional Neural Networks (CNN)~\cite{gilles2021home,multimodal} are applied to encode features from scenes and agents. 
Then, Graph Neural Networks~\cite{graph1,graph2,graph3} are used to further model the interaction between agents. 
With the innovation of the Transformer structure in many research fields, many works have applied it to multimodal trajectory prediction~\cite{sceneformer,motionformer}. 
However, these efforts are troubled by the inherent coupling issue of the trajectory prediction task, that is, the mutual influence between multiple intelligent agent trajectories. 
To address this, GameFormer~\cite{gameformer} proposes a hierarchical trajectory refinement formulated by \textit{level-k} game theory.
It decouples interactions between agents within the same game level by predicting agent trajectories in each level based on other agents from the last game level.
While GameFormer achieves cutting-edge performance, it involves redundancy in its \textit{level-k} game refinement, as every agents are treated equally in all game levels.
Due to varying environmental conditions, different agents encounter driving environments of differing complexities. This implies that the required depth of game varies for each distinct agent.
Thus, we combine the multimodal trajectory prediction task with the \textit{level-k} game theory, providing a measurement for game state stability for agents.
By adding a simple mask module, we effectively improve the performance of GameFormer, and reducing its computational cost at the same time.

\subsection{Joint Trajectory Prediction and Planning}
The ultimate goal of trajectory prediction is to enable autonomous vehicle to make safe and efficient decisions in complex driving environments~\cite{lookout,deepim}. The integration between trajectory prediction and planning poses a critical challenge in autonomous driving, as the trajectories of agents interact with each other~\cite{dipp}. Some approaches establish a sequential connection between planning and prediction~\cite{pip,mp3}, focusing on a single vehicle and querying the trajectory prediction models of other vehicles. However, achieving reliable trajectories necessitates dense computations in this framework, significantly constraining computational speed~\cite{dtpp}.

Another promising paradigm involves joint trajectory prediction and planning. It utilizes a single model to implicitly consider interactions among multiple agents, providing end-to-end output that is globally consistent~\cite{dipp,dtpp,sceneformer}. Nevertheless, this paradigm often lacks interpretability and struggles to yield reasonable results in complex scenarios.
To enhance the interpretability of joint prediction and planning, Gameformer~\cite{gameformer} introduces K-layer game theory, explicitly modeling interactions among agents. Leveraging a K-layer Transformer Decoder, GameFormer notably enhances the accuracy of trajectory prediction and planning. Our work build upon the K-layer game theory of GameFormer, and further refine it. By modeling the game states of agents, we eliminate redundancies in the K-layer game, leading to enhanced precision and computational efficiency in joint trajectory prediction and planning.

\section{Methodology}
In this section, we first describe the preliminaries of our method, including the description of the Multimodal Trajectory Prediction (MTP) task in \cref{sec:MTP} and the details of our baseline, GameFormer, in \cref{sec:GF}. Then, we elaborate the proposed Trajectory Entropy in \cref{sec:TE}. Finally, the improved level-k game framework for joint trajectory prediction and planning is presented in \cref{sec:IG}.

\subsection{Preliminaries}\label{sec:pre}

\subsubsection{Multimodal Trajectory Prediction}\label{sec:MTP}
Trajectory prediction is a pivotal module in autonomous driving systems. Initially, trajectory prediction was formulated to predict a single future trajectory $\mathbf{y}_{i}$ of an agent $A_{i}$ from input $X_{i}$ using a model $\mathcal{F}$, known as discriminative trajectory prediction.
\begin{equation}
	\mathbf{y}_{i}=\mathcal{F}(X_{i}).
\end{equation}
Here, the trajectory $\mathbf{y}_{i}$ represents a sequence of states (i.e., 2D coordinates sequence) over the future horizon $T$.
However, due to the inherent uncertainty in agent trajectories, discriminative trajectory prediction fails to encompass all feasible solutions. This limitation hinders the network from learning reasonable driving policy. To address this issue, Multimodal Trajectory Prediction (MTP) tasks have been introduced~\cite{socialgan}.
In this task, the model predicts a distribution covering all feasible solutions by outputting multiple trajectories with confidences for an agent, thus handling the uncertainty in trajectories.

\begin{equation}
	\mathbf{Y}_{i} = \mathcal{F}(X_{i}).
\end{equation}
Here, $\mathbf{Y}_{i}=\{(\mathbf{y}_{j},c_j)\}_j^M$, where $\mathbf{y}_{j}$ is an individual trajectory, $c_{j}$ is corresponding confidence, and $M$ is the number of modalities.
Due to enhanced interpretability and performance, this approach has become the default configuration for current trajectory prediction tasks~\cite{MTP_survey}.

\begin{figure}[!t]
\centering
\includegraphics[width=\linewidth]{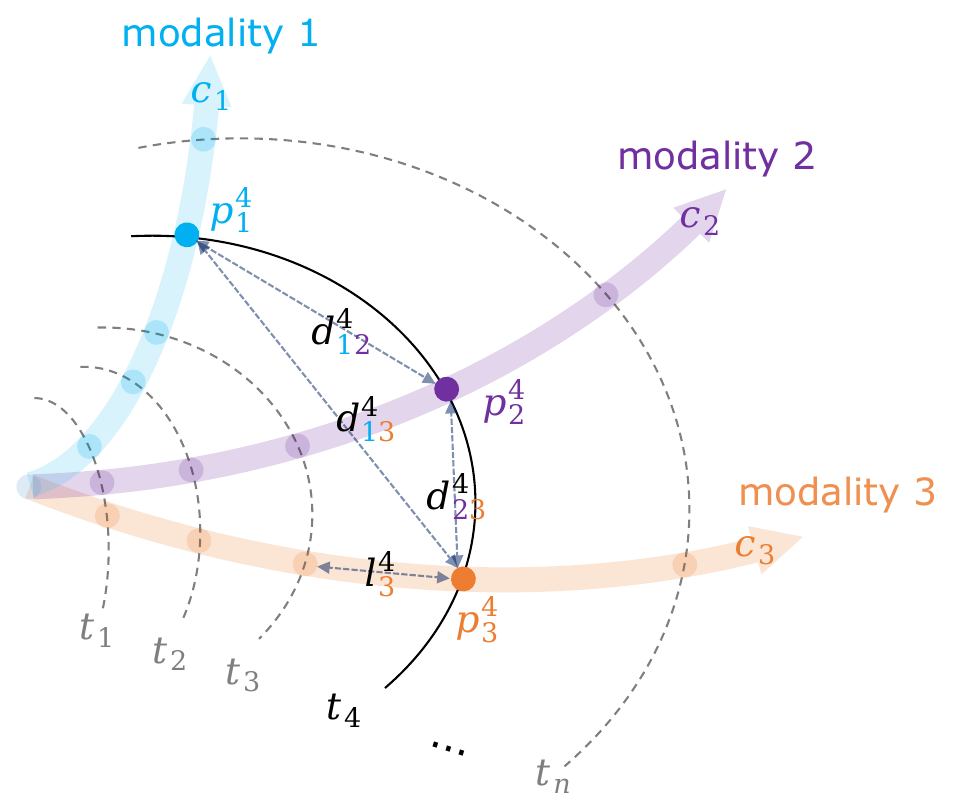}
\caption{\textbf{The formulation of Trajectory Entropy from Multimodality Trajectory Prediction (MTP).} Our Trajectory Entropy essentially extracts statistical signals indicating uncertainty from MTP, utilizing the trajectory confidences ($\{c_i\}$), the distances of trajectories between modalities ($\{d_{i,j}^t\}$) and the trajectory length within unit time interval ($\{l_i^t\}$).}
\vspace{-1.6em}
\label{fig:te-theory}
\end{figure}

\subsubsection{GameFormer}\label{sec:GF}
The joint trajectory prediction and planning contributes to enhancing the intelligent decision-making capability of the learning model. However, a primary challenge of this approach lies in the strong coupling of	 trajectories among agents. That is the agents mutually influence the decisions of each other.
To address this issue, GameFormer~\cite{gameformer} explicitly models interactions among agents through the level-k game theory~\cite{gamet}. Specifically, it employs a Decoding $\mathcal{D}$ to decode trajectories for the ego vehicle and other vehicles. This module consists of $K$ layers of decoders, predominantly based on Transformer networks. For any agent $A_i$,  the $k$-th decoder uses the outputs of the $k\!-\!1$-th decoder as the input, which includes trajectories of all other agents $\{\mathbf{Y}_{j}^{k-1}\}_{j\not=i}$, to decode the trajectory of $A_i$ in the layer of the game, except for the $0$-th layer.

\begin{equation}
	\mathbf{Y}_{i}^{k} = \mathcal{D}(\{\mathbf{Y}_{j}^{k-1}\}_{j\not=i},\mathbf{C}_{s}),~k\in(0,K]
\end{equation}
where $\mathbf{C}_{s}$ represents scene features obtained by the Transformer encoder. Through this approach, GameFormer decouples interactions among agents in a hierarchical game, enhancing interpretability and the overall accuracy.

However, treating all agents equally in the k-level games can introduce noise and  computational redundancies. Due to varying environmental conditions, driving difficulty faced by different agents is certainly different. Some agents with lower driving difficulty can achieve stable results in the shallow-level games, while others with greater difficulty require deeper reasoning. Therefore, it is reasonable to fix agents with lower driving difficulty in early game layers, as their trajectory prediction results are already stable. This not only reduces computational costs but also decreases overall uncertainty in the game, thereby improving accuracy. Hence, by analyzing the inherent uncertainty nature of multimodal trajectory prediction, this paper proposes a measurement to reveal the game state stability of agents (\cref{sec:TE}). Then, the level-k game framework of GameFormer can be enhanced in terms of both accuracy and efficiency, by a simple gate module (\cref{sec:IG}).

\begin{figure}[!t]
\centering
\includegraphics[width=\linewidth]{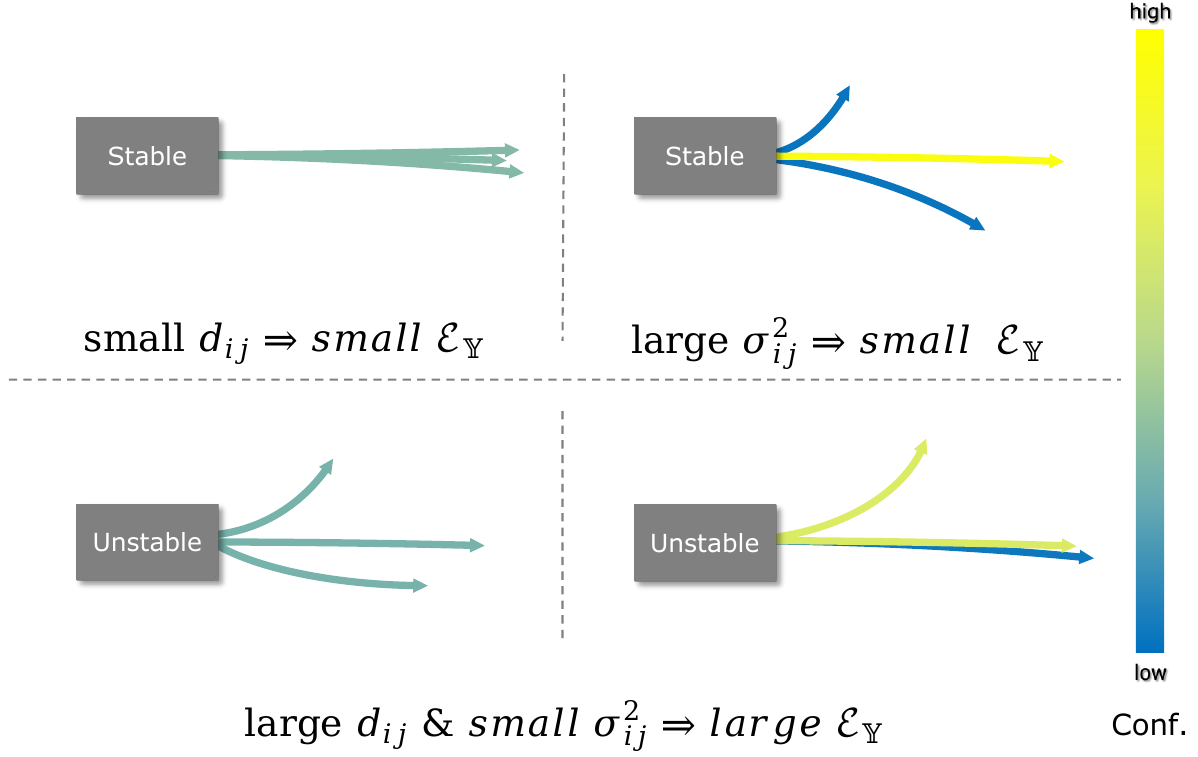}
\caption{\textbf{Some toy examples of Trajectory Entropy. Top:} A low Trajectory Entropy may arise from concentrated trajectories (left) or a single trajectory with high confidence (right) in MTP. Both of them exhibits narrow distribution of feasible trajectories. \textbf{Bottom:} Conversely, in cases where MTP comprises diverse trajectories with comparable confidence levels, the corresponding Trajectory Entropy is high.}
\vspace{-1.6em}
\label{fig:te-exp}
\end{figure}

\subsection{Trajectory Entropy}\label{sec:TE}
To improve performance of level-k game framework for AD, we utilize the uncertainty intrinsic of the Multimodal Trajectory Prediction (MTP) to formulate Trajectory Entropy (cf. \cref{fig:te-theory}). 
This metric, then, can be used to measure the stability of agent state in the level-k game, enabling adaptively assignment of proper reasoning depth for different agents.
This ultimately enhances the performance of trajectory prediction and planning. 
Next, we will provide a detailed explanation of the Trajectory Entropy.

The MTP results of any agent $A$ can be expressed as: $\mathbf{Y}=\{(y_{j},c_{j})\}_{j}^{M}$, where $M$ is the number of modalities.
The $y_{j}=\{p_{j}^{t}\}_{t}^{T}$ represents the predicted trajectory of the $j$-th modality, comprising discrete 2D trajectory points $\{p_{j}^{t}=(u_{j}^{t},v_{j}^{t})\}_{t}^{T}$ over the future horizon $T$.
The $c_{j}$ denotes the confidence of the $j$-th modality trajectory.

For notational simplicity, we use a symbolic representation of Trajectory Entropy: $\mathcal{E}_{\mathbf{Y}}$.
This metric measures the uncertainty of the MTP results: $\mathbf{Y}=\{(y_{j},c_{j})\}_{j}^{M}$. It can reflect the game states of agents in the level-k game framework, which essentially is a clue of agent driving difficulty.

Next, We simplify this temporal uncertainty measurement onto specific time step $t$. The trajectories are composed of discrete trajectory points in time. Hence, without loss of generality, we assume that the overall uncertainty $\mathcal{E}_{\mathbf{Y}}$ is the accumulation of uncertainties of trajectory point set $\mathcal{E}_{\mathbf{P}_{t}}$ at each time step $t$, where $\mathbf{P}_{t} = \{p^{t}_{j}, c_{j}\}_j^{M}$ is a trajectory point set at a time step $t$ (cf. \cref{fig:te-theory}).
\begin{equation}
	\mathcal{E}_{\mathbf{Y}} = \sum_{t}^{T} \mathcal{E}_{\mathbf{P}_{t}}.
\end{equation}
In other words, we assume this stochastic process is 0-dependent~\cite{prs}, with the confidence at each time step being the same as the overall confidence.
This assumption may not be flawless, due to the interrelated nature of trajectory points in a time series.
However, the uncertainty in MTP is predominantly influenced by the interactions among trajectories rather than by changes within a single trajectory over time. 
Thus, our assumption proves effective on uncertainty measurement in our experiments, showcasing impressive performance.

Then, we focus on the uncertainty of the trajectory point set $\mathcal{E}_{\mathbf{P}_{t}}$ at a specified time step $t$.
Intuitively, the uncertainty of a multimodal trajectory set is related to \textit{the distances between trajectories} and \textit{the confidence levels of each trajectory}. We utilize them to form the core of Trajectory Entropy. 

Firstly, we calculate the Euclidean distances of point pairs ($d^t_{ij}=||p^t_i - p^t_j||_2$) from M-modality trajectories: $\{d^{t}_{ij}\}_{i\not=j\in M}$.
Without considering the confidences, the greater the distance between trajectory points, the more dispersed the points are, leading to higher uncertainty in trajectory prediction. Therefore, this distance is proportional to the uncertainty of the point set, serving as an important factor of $\mathcal{E}_{\mathbf{P}_{t}}$.

Taking a step further, we then incorporate the trajectory confidences $\{c_{i}\}_i$ into the calculation of $\mathcal{E}_{\mathbf{P}_{t}}$.
 Drawing inspiration from random signal processing theory~\cite{rsp}, $\{d^{t}_{ij}\}_{i\not=j\in M}$ can be viewed as signals reflecting the uncertainty of the trajectory point set. 
We assume each of these signals includes a white Gaussian noise~\cite{rsp} with a zero mean, whose variance is formed through the confidences of the trajectory distances: $\{\sigma_{ij}^{2}=1/c_{i}c_{j}\}_{i\not=j\in M}$. 
The intuitive principle is that the confidences are \textbf{inversely} proportional to the level of noise (\ie, the noise power $P^{noise}$) in the prediction trajectories.
Because each distance consists of two trajectories with confidences $c_{i}$ and $c_{j}$, the power of noise within each distance signal is :
 \begin{equation}
 	P^{noise}_{ij} = \sigma_{ij}^{2}=1/c_{i}c_{j}.
 \end{equation}
Similarly, the power of each distance signal self is:
 \begin{equation}
 	P^{signal}_{ij} = (d_{ij}^{t})^2.
 \end{equation}
Therefore, the signal-to-noise ratio (SNR) of each distance signal can be expressed as: 
\begin{equation}
	{SNR}^{t}_{ij}=d_{ij}^{t2}/\sigma_{ij}^{2}.
\end{equation}
These distance signals constitute the overall uncertainty of the trajectory point set $\mathbf{P}_{t}$ and the SNRs quantitatively reflect the strength of these signals under the prediction noises.
Thus, the overall uncertainty of the trajectory point set $\mathbf{P}_{t}$ can be represented as the sum of these SNRs:
\begin{equation}
	\mathcal{E}_{\mathbf{P}_{t}}=\sum\limits_{i,j} SNR^{t}_{ij}.
\end{equation}
By accumulating it over time $t$ with a normalization factor, we can obtain the MTP uncertainty, \ie \textit{Trajectory Entropy}.
\begin{equation}\label{eq:et-f}
	\mathcal{E}_{\mathbf{Y}} =  \sum_{t}^{T} \frac{\mathcal{E}_{\mathbf{P}_t}}{E^{t}_{M}(\mathbf{l})}.
\end{equation}
Here, $E^{t}_{M}(\mathbf{l})=\sum\limits_{j}^{M}c_{j}l^{t}_{j}$ denotes the expectation of distance traveled by multimodality trajectories within a unit time duration, essentially representing the expectation of instantaneous speed, where $l^t_j = ||p_j^{t}-p_j^{t-1}||^2_2$ (cf. \cref{fig:te-theory}). 
Variations in speed result in changes in the spatial scales of different trajectory points.
However, the alterations of spatial scale have been encoded in the distance signals.
Thus, $E^{t}_{M}(\mathbf{l})$ serves as a normalization factor to remove the effect of speed variations across different time steps.

As the SNR increases, more effective information emerges from the uncertainty signals~\cite{rsp}. Following the definition of information entropy in information theory~\cite{IT}, higher information content corresponds to higher entropy and thus greater uncertainty, aligning with our perspective on describing MTP ($\mathbf{Y}$) uncertainty . Therefore, we refer to this measurement constructed by SNRs of uncertainty signals as \textit{Trajectory Entropy}, denoted as $\mathcal{E}_{\mathbf{Y}}$. We provide some toy examples in \cref{fig:te-exp}, showing that the \textit{Trajectory Entropy} is in line with the human intuition.

\begin{figure}[!t]
\centering
\includegraphics[width=\linewidth]{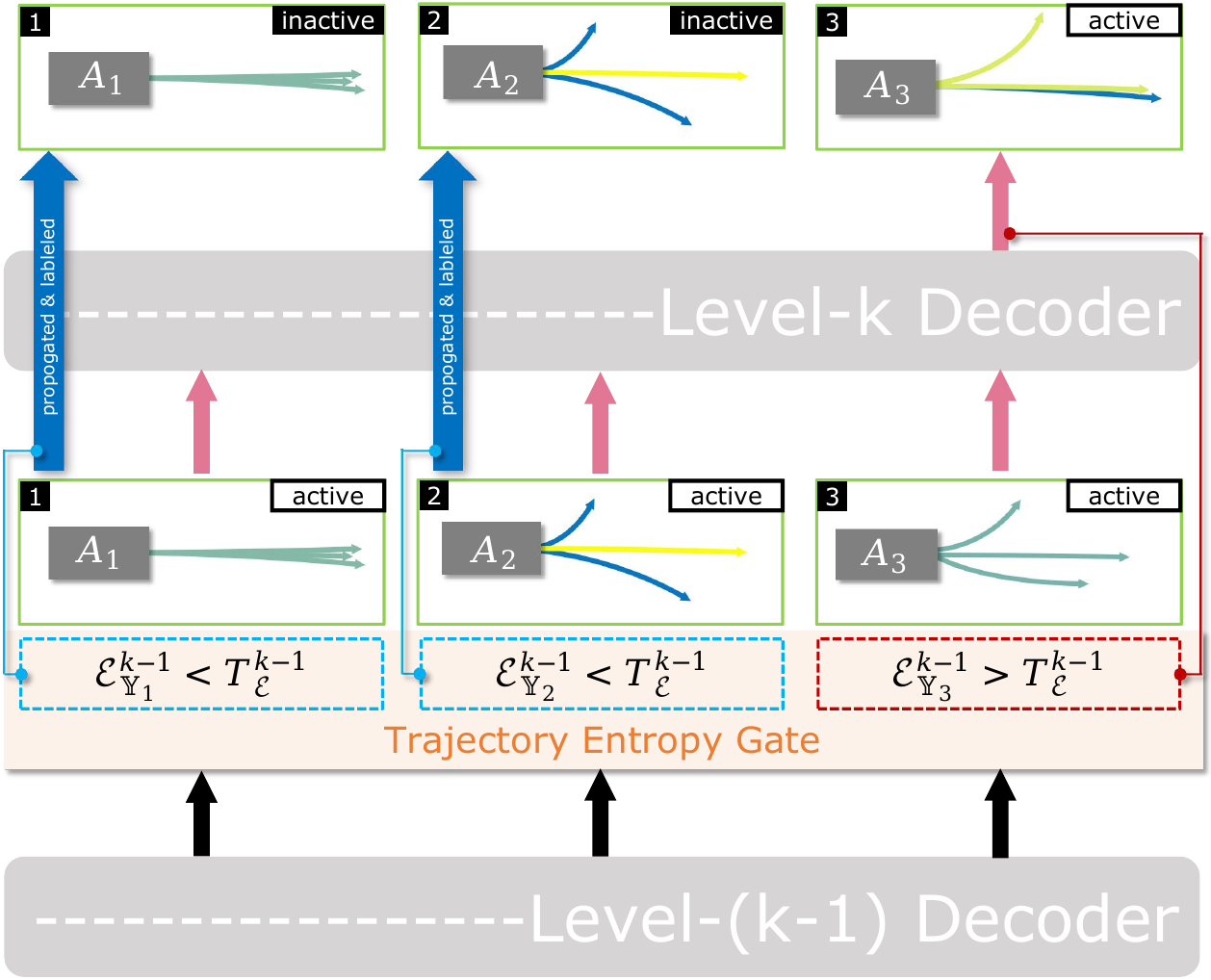}
\caption{\textbf{The hierarchical game framework improved by Trajectory Entropy.} Before the level-(k-1) decoder, a Trajectory Entropy Gate is set up with a specified threshold ($T_{\mathcal{E}}$), where the MTP results of active agents ($\mathbf{Y}_i$) are utilized to calculate the Trajectory Entropy ($\mathcal{E}_{\mathbf{Y}_i}$). If the $\mathcal{E}_{\mathbf{Y}_i} < T_{\mathcal{E}}$, the agent $A_i$ is labeled as inactive. The MTP results of all the inactive agents are directly propagated to the subsequent game levels, while those of active agents are processed by the level-k decoder.}
\vspace{-1.6em}
\label{fig:ge-imp}
\end{figure}

%
%
%

\subsection{Improving level-k Game Framework}\label{sec:IG}
Through Trajectory Entropy, we can utilize the uncertainty inherent in MTP to assess the game states of agents, thereby enhancing the level-k game framework of GameFormer. 
Specifically, low trajectory entropy indicates that model prediction for the feasible trajectory distribution of an agent is highly concentrated. In driving scenarios with structured rules, this often implies that agent faces low driving difficulty. Therefore, within the level-k game, such agents could achieve stable game results in current game level and does not require further deep reasoning. We can then fix their MTP results in order to avoid redundant computations.

This enhancement is implemented by a simple Trajectory Entropy Gate in \cref{fig:ge-imp}, positioned before all level decoders except for the $0$-th level.
 The function of the Gate is to classify agents as active or inactive based on their Trajectory Entropy. 
Taking the $k$-th level as an example, for all active agents $\{A_i\}_i$, we only need to calculate the trajectory entropy of their MTP results before the $k$-th level decoder (\ie, the results of $k-1$-th level decoder): $\{\mathcal{E}^{k-1}_{\mathbf{Y}_i}\}_i$ and compare with a manually set threshold $T^{k-1}_{\mathcal{E}}$. If $\mathcal{E}^{k-1}_{\mathbf{Y}_i} < T^{k-1}_{\mathcal{E}}$, the MTP result $\mathbf{Y}_i$ of the corresponding agent is directly propagated to all the subsequent game levels, and the agent is labeled as inactive. 
Thus, the computational redundancies can be reduced and the overall accuracy can be improved as well, as shown in our experiments.
It is worth noting that the Trajectory Entropy threshold can vary across different game levels. 
The intuition behind this setting is that the high game levels tend to exhibit low entropy in the MTP results (cf. \cref{fig:te-ana}), due to their deeper reasoning.
Hence, the threshold $\{T^k_{\mathcal{E}}\}_k$ set for the higher game levels is adjusted to a lower value, reflecting a more stringent criterion for labeling agents as inactive.

\section{Experiments}
In this section, we evaluate the performance of the proposed \textit{Trajectory Entropy} in terms of improving level-k game framework on the joint trajectory prediction and planning task. Firstly, we present the implementation details of our method in~\cref{sec:id}. Two benchmarks are involved in our experiments, \ie, the Waymo Open Motion Dataset (WOMD)~\cite{waymo} and nuPlan~\cite{nuplan} datasets, both of which are utilized to conducted trajectory prediction (cf. \cref{sec:tpr}) and ego planning~(cf.~\cref{sec:pr}) experiments.
Further ablation studies can be found in \cref{sec:ab}. 

\begin{table}[!t]
\caption{Comparison with state-of-the-art models on the WOMD interaction prediction benchmark. The best results are \textbf{highlighted}.}
\centering
\resizebox{0.9\linewidth}{!}{%
\begin{tabular}{@{}lcccc@{}}
\toprule
Model                                   & minADE ($\downarrow$) & minFDE ($\downarrow$) & Miss rate ($\downarrow$) & mAP ($\uparrow$) \\ \midrule
LSTM baseline \cite{waymo}  & 1.9056                &  5.0278               &  0.7750                  & 0.0524    \\
Heat \cite{mo2022multi}                 & 1.4197                &  3.2595               &  0.7224                  & 0.0844    \\
AIR$^2$ \cite{wu2021air}                & 1.3165                &  2.7138               &  0.6230                  & 0.0963    \\
SceneTrans \cite{sceneformer}        & 0.9774                &  2.1892               &  0.4942                  & 0.1192     \\
DenseTNT \cite{gu2021densetnt}          & 1.1417                &  2.4904               &  0.5350                  & 0.1647   \\
M2I \cite{sun2022m2i}                   & 1.3506                &  2.8325               &  0.5538                  & 0.1239     \\
MTR \cite{mtr}                & 0.9181                & 2.0633                &  \textbf{0.4411}         & \textbf{0.2037} \\ \midrule
GameFormer~\cite{gameformer}                   & {0.9177}       & {1.9791}       &  0.4747                  & 0.1211 \\
\ours Ours                   &\ours  \textbf{0.9092}       &\ours  \textbf{1.9673}       &\ours   0.4643       &\ours  0.1226 \\ 
 \bottomrule
\end{tabular}
}
\vspace{-0.2cm}
\label{tab:waymo-pred}
\end{table}

\subsection{Experimental Details}\label{sec:id}
\subsubsection{Model Details}
Our method utilizes GameFormer as the backbone, trained on two datasets separately. Next, we present the implementation details of the model on two datasets separately, including its training and inference settings.

\paragraph{For Waymo Open Motion Dataset (WOMD):}
Following~\cite{gameformer}, we train and evaluate the original GameFormer and our improved version on the Waymo open motion dataset, in terms of interaction prediction, open-loop and closed-loop planning.
Specifically, for the interaction prediction task, we train two models (original GameFormer and ours) on the entire Waymo open motion dataset following~\cite{gameformer}. The joint prediction setting is applied for this task, where only 6 joint trajectories of the two agents are predicted~\cite{sceneformer}.
For the planning tasks, we use the scenarios provided by GameFormer~\cite{gameformer}, including 10K 20-second randomly selected scenarios. 9K scenarios are used for training and the remaining 1K for validation, following GameFormer~\cite{gameformer}. The models are trained on 4 NVIDIA A100 GPU, utilizing the official implementation.
Then, the joint prediction and planning performance is evaluated in 400 8-second interactive and dynamic scenarios.
The GameFormer model in this dataset is set with $5$ decoder layers (level-$0/1/2/3/4$) for planning and $4$ decoder layers (level-$0/1/2/3$) for prediction, according to~\cite{gameformer}.
The Trajectory Entropy thresholds of our improved version are set differently in specific experiments.

\begin{table}[!t]
\caption{Trajectory prediction results on the nuPlan benchmark. The error reduction of our method compared with the original GameFormer is showcased as {\imp subscripts}.}
\centering
\resizebox{\linewidth}{!}{
\begin{tabular}{lllll}
\toprule
Method            & minADE~($\downarrow$)     & minFDE~($\downarrow$)        & Miss Rate~($\downarrow$)   & mAP~($\uparrow$)    \\ \midrule

GameFormer~\cite{gameformer}        & 0.4728      & 0.8510    & 0.0186         & 0.2064                     \\

\ours Ours        & \ours{0.4297}{\imp$_{-9.12\%}$}          & \ours{0.6876}{\imp$_{-19.20\%}$}    & \ours{0.0149}{\imp$_{-19.89\%}$}           & \ours{0.2118}{\imp$_{+2.62\%}$}                    \\
\bottomrule
\end{tabular}%
}
 \label{tab:nuplan-pred}
\vspace{-0.2cm}
\end{table}

\paragraph{For nuPlan:} 
nuPlan stands as the premier large-scale benchmark for autonomous driving planning, consisting of massive driving scenes and data from real world.
To train the GameFormer model, we evenly sample 60K scene datas from the whole nuPlan training dataset. 
Then, 50K scenes are allocated for model training, while the remaining 10K scenes are reserved for trajectory prediction evaluation. The planning evaluation, on the other hand, is performed on the selected test set of nuPlan, following~\cite{plantf}, which is a standard benchmark.
The GameFormer model is implemented with 3 decoder layers (level-$0/1/2$). 
The training and inference are performed on 1 GeForce RTX 4090 GPUs, following the hyper-parameter setting of the official implementation. 
During inference, the Trajectory Entropy thresholds are set according to specific driving difficulties. 
In practice, the thresholds are set differently in different scenes, as the scene difficulty varies in the nuPlan dataset.
The specific thresholds are described in the following experimental sections.
To achieve the single planning trajectory from MTP results, we use a trajectory selection approach proposed by~\cite{pdm}, involving a score function for trajectory selection. 

\subsubsection{Evaluation Metrics}
In the following, we describe the evaluation metrics for prediction and planning.

\paragraph{Trajectory Prediction Metrics:}
For both WOMD and nuPlan benchmarks, we use the standard evaluation metrics~\cite{gameformer}, including minimum average displacement error (minADE), minimum final displacement error (minFDE), miss rate, and mean Average Precision (mAP).

\paragraph{Planning Metrics:}
In open-loop planning of WOMD datasets, following~\cite{gameformer}, we use distance-based error metrics to verify the planning performance, including collision rate, miss rate, prediction and planning error (\ie ADE and FDE).
In closed-loop planning of WOMD datasets, we also following~\cite{gameformer}, using metrics consisting of including success rate, progress along the route, comfort-related metrics (\ie longitudinal acceleration and jerk, lateral acceleration), and position errors.
In nuPlan benchmark, we conduct planning experiments under various settings, comprising open-loop and closed-loop with reactive/nonreactive agents. A comprehensive planning score is applied to performance evaluation in these settings, which is provided by the nuPlan platform~\footnote{https://github.com/motional/nuplan-devkit}.

\subsection{Trajectory Prediction Results}\label{sec:tpr}

\subsubsection{Waymo Interaction Trajectory Prediction}

Firstly, we conduct the interaction prediction experiment on the WOMD, using the official setting~\cite{waymo,gameformer}.
Our model predicts the joint future positions for the two interacting agents over an 8-second time horizon.
The comparison methods include recent \textit{state-of-the-art} approaches~\cite{waymo,mo2022multi,wu2021air,sceneformer,sceneformer,mtr,gameformer,sun2022m2i}.
For the experiment, our Trajectory Entropy thresholds are set as $[T^{0}_{\mathcal{E}}\!=\!4.3,~T^{1}_{\mathcal{E}}\!=\!4.2,~T^{2}_{\mathcal{E}}\!=\!4.1]$.
We report the prediction results in \cref{tab:waymo-pred}.
Our method demonstrates improved accuracy over the original GameFormer across \textbf{all} metrics, validating the effectiveness of the proposed Trajectory Entropy. Additionally, it achieves the best minADE and minFDE scores among all approaches, further highlighting its superior performance.

\subsubsection{nuPlan Trajectory Prediction}
Although the nuPlan benchmark is mainly used for planning evaluation, we also conduct trajectory prediction experiments on it.
We utilize models to predict the positions of neighboring agents 8 seconds into the future. 
In this experiment, the Trajectory Entropy thresholds are set as $[T^{0}_{\mathcal{E}}\!=\!40,~T^{1}_{\mathcal{E}}\!=\!30]$.
Our method results are compared with the original GameFormer. 
The results are summarized in \cref{tab:nuplan-pred}. 
As we can see that, Trajectory Entropy Gate improves the accuracy of GameFormer in terms of all metrics. 
The prediction error is reduced by a large margin, \eg, up to $19.89\%$.
This proves that properly assigning different reasoning depths to different agents is highly beneficial for the overall prediction precision.
Our \textit{Trajectory Entropy} significantly enhances the performance of the level-k game framework, paving the way to intelligent autonomous driving.

\begin{figure*}[!t]
\centering
\includegraphics[width=\linewidth]{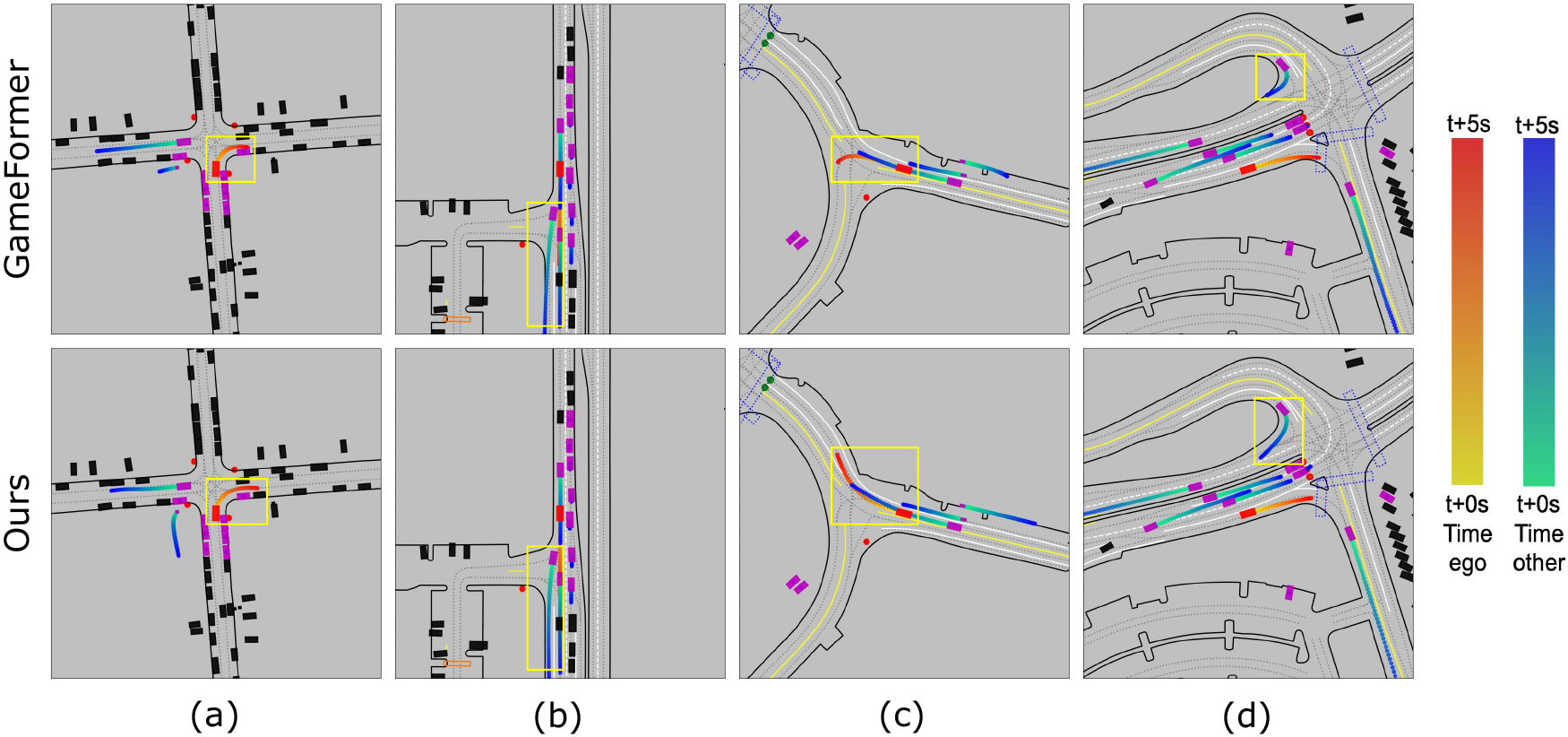}
\caption{\textbf{The qualitative comparison results between our method with GameFormer in the open-loop planning task of WOMD dataset.} It can be seen that our method provides more reasonable trajectories of both ego and other agents. Specifically, our method enables corruption avoidance in (a), ensures trajectory predictions remain within drivable areas in (b) and (d), and facilitates correct direction planning in (c). The red trajectory is the ego agent plan and the blue ones are the predictions of neighboring agents.}
\vspace{-1.6em}
\label{fig:qua}
\end{figure*}

\begin{table*}[!t]
\caption{Evaluation of open-loop planning performance in selected WOMD scenarios. The best results are \textbf{highlighted}.}\label{tab:waymo-ol}
\centering
\small
\begin{tabular}{@{}lccccc}
\toprule
\multirow{2}{*}{Method} &
  \multirow{2}{*}{Collision rate $\downarrow$ (\%)} &
  \multirow{2}{*}{Miss rate $\downarrow$ (\%)} &
  \multicolumn{3}{c}{Planning error $\downarrow$ (m)}     \\

                &               &                 & @1s            & @3s            & @5s          \\  \cmidrule(r){1-1} \cmidrule(r){2-2} \cmidrule(r){3-3} \cmidrule(r){4-6}
DIPP~\cite{dipp}           & 4.45          & \textbf{7.71}             & 0.276          & 1.236          & 3.282               \\
GameFormer~\cite{gameformer}           & {4.39} & {12.11}    & \textbf{0.083} &{0.877}  & {2.685} \\
 \ours Ours           &  \ours \textbf{2.65} & \ours  {9.75}    &  \ours {0.087} & \ours \textbf{0.832}  &  \ours \textbf{2.513}  \\ \bottomrule
\end{tabular}
\vspace{-0.4cm}
\end{table*}

\subsection{Planning Results}\label{sec:pr}

\subsubsection{Waymo Open-loop Planning}
We conduct the open-loop planning experiment on the WOMD.
Following~\cite{gameformer}, selected WOMD scenarios with a prediction/planning horizon of 5 seconds are employed in our experiments. 
We perform $M=6$ MTP for 10 neighboring agents closest to the ego vehicle.
A recent \textit{state-of-the-art} planning method based on imitation learning is set as the baseline~\cite{dipp}, along with the original GameFormer~\cite{gameformer}.
Our Trajectory Entropy thresholds are set as $[T^{0}_{\mathcal{E}}\!=\!0.8,~T^{1}_{\mathcal{E}}\!=\!0.75,~T^{2}_{\mathcal{E}}\!=\!0.7,~T^{3}_{\mathcal{E}}\!=\!0.65]$ in this experiment.
The results are reported in~\cref{tab:waymo-ol}. 
It can be seen that our method achieves better results than the original GameFormer. Especially, a substantial decrease in collision rate is observed.
Overall, the lower planning errors also demonstrate the superiority of our method against current \textit{state-of-the-art}s~\cite{dipp,gameformer}.
We also provide some visualization comparison between our method and GameFormer in \cref{fig:qua}.
In the figure, the trajectories from our method are more reasonable than those from GameFormer, proving that our Trajectory Entropy is able to improve the performance of the level-k game framework.

\subsubsection{Waymo Closed-loop Planning}

			\begin{table*}[htp]
\caption{Evaluation of closed-loop planning performance in selected WOMD scenarios. The \textbf{best} results are highlighted.}
\centering
\resizebox{0.95\linewidth}{!}{
\begin{tabular}{lcccccccc}
\toprule
\multirow{2}{*}{Method} &
\multicolumn{1}{c}{Success rate$~\uparrow$} &
\multicolumn{1}{c}{Progress$~\uparrow$} &
\multicolumn{1}{c}{Acceleration$~\downarrow$} &
\multicolumn{1}{c}{Jerk$~\downarrow$} &
\multicolumn{1}{c}{Lateral acc.$~\downarrow$} &
\multicolumn{3}{c}{Position error$~\downarrow$ ($m$)} \\
                    &  (\%)                 & $(m)$                 & ($m/s^2$)             & ($m/s^3$)             & ($m/s^2$)             & @3s                   & @5s            & @8s \\ \cmidrule(r){1-1} \cmidrule(r){2-3} \cmidrule(r){4-6} \cmidrule(r){7-9}
DIPP~\cite{dipp}                & 85.43        & 45.04       & 0.76         & 3.42       & 1.48    & 2.81       & 6.09 & 12.65\\
GameFormer~\cite{gameformer}                & 82.25      & 51.25       & \textbf{0.54}  & 3.80	      & 2.44        & 2.20        & 4.74 & 9.25 \\

%

  \ours Ours                &   \ours \textbf{87.25}        & \ours   \textbf{53.31}        & \ours   0.65         & \ours   \textbf{1.42}        & \ours   \textbf{0.26}         & \ours   \textbf{2.18}         & \ours   \textbf{4.64} & \ours   \textbf{8.47} \\

\bottomrule
\end{tabular}
}
\vspace{-0.3cm}
\label{tab:way-cl}
\end{table*}

The closed-loop planning performance is also evaluated on the WOMD dataset.
We use the simulation environment provided by~\cite{dipp}, where the state of ego agent is updated by planning model and other agents follow logged trajectories in the dataset. 
The test dataset selected from WOMD dataset is provided by DIPP~\cite{dipp}, also recommended by GameFormer~\cite{gameformer}.
Our method is compared with these recent \textit{state-of-the-arts}~\cite{dipp,gameformer}.
The results are reported in \cref{tab:way-cl}. 
It can be seen that our method surpasses other approaches in terms of most metrics, achieving the best success rate and the smallest position error.
This indicates our Trajectory Entropy effectively improves the planning performance, providing safer and more comfortable driving trajectories.

\subsubsection{nuPlan Planning Benchmark}
The nuPlan benchmark is employed to assess the planning performance as well. 
We use the standard test scenes provided by~\cite{plantf}, including \textbf{Test14-random} and \textbf{Test14-hard}.
The former is sampled randomly from 14 scenario types and the latter consists of hard scenarios selected by PDM~\cite{pdm}.
The Trajectory Entropy thresholds are set as $[T^{0}_{\mathcal{E}}\!=\!30,~T^{1}_{\mathcal{E}}\!=\!28]$ as default.
In OLC and NR-CLS of Test14-hard, we slightly decrease $T^{1}_{\mathcal{E}}$ to $25$ to handle more challenging scenes.
We compare our methods with rule-based~\cite{idm,pdm}, learning-based~\cite{nuplan,urban,pgp,pdm,plantf,pluto} and hybrid palnning method~\cite{pdm,gameformer,pluto}. Note the hybrid methods involve a trajectory refinement utilizing rule-based strategies. Our method belongs to this category, same as the original GameFormer.
The results are summarized in \cref{tab:nuplan}.
It can be seen that our improved GameFormer outperforms the original version by a large margin on all scenes, especially in reactive closed-loop planning scenes, \eg, up to $16.48\%$ ($79.31 \rightarrow 92.38$).
This indicates that the level-k game framework can produce more reasonable results when agents with different driving difficulties are distinguished and assigned with different reasoning depths.
Meanwhile, our method achieves comparable or better results than the recent cutting-edge planning methods~\cite{pluto,pdm,zheng2025diffusionbased}, implying the potential of the game theory-based framework in autonomous driving planning.
It is also worth-noting that our method is able to perform trajectory prediction and planning at the same time, which demonstrates the superiority of our method than planning-only methods, such as PLUTO~\cite{pluto} and Diff. Plan~\cite{zheng2025diffusionbased}.

\begin{table*}[!t]
\vspace{6pt}
\begin{center}
\caption{Comparison with state-of-the-arts on the nuPlan planning benchmark. The \textbf{best} and \underline{second} results in \textbf{\textit{hybrid series}} are {highlighted}. 
}
\label{tab:nuplan}
\setlength{\tabcolsep}{10pt}
\renewcommand{\arraystretch}{1.2}
\small
\begin{tabular}{ccllllll}
\toprule
\multicolumn{2}{c}{Planners} & \multicolumn{3}{c}{Test14-random} & \multicolumn{3}{c}{Test14-hard}  \\ \cmidrule(r){1-2} \cmidrule(r){3-5} \cmidrule(r){6-8}
Type & \multicolumn{1}{l}{Method} & OLS & NR-CLS & \multicolumn{1}{l}{R-CLS} & OLS & NR-CLS & \multicolumn{1}{l}{R-CLS}  \\ \cmidrule(r){1-2} \cmidrule(r){3-5} \cmidrule(r){6-8}

\fadedtext{Expert} & \multicolumn{1}{l}{\fadedtext{Log-replay}} & \fadedtext{100.0} & \fadedtext{94.03} & \multicolumn{1}{l}{\fadedtext{75.86}} & \fadedtext{100.0} & \fadedtext{85.96} & \multicolumn{1}{l}{\fadedtext{68.80}} \\ \cmidrule(r){1-2} \cmidrule(r){3-5} \cmidrule(r){6-8}
\multirow{2}{*}{Rule-based} & \multicolumn{1}{l}{IDM~\cite{idm}} & 34.15 & 70.39 & \multicolumn{1}{l}{72.42} & 20.07 & 56.16 & \multicolumn{1}{l}{62.26} \\
 & \multicolumn{1}{l}{PDM-Closed~\cite{pdm}} & {46.32} & {90.05} & \multicolumn{1}{l}{{{91.64}}} & {26.43} & {65.07} & \multicolumn{1}{l}{{75.18}} \\ \cmidrule(r){1-2} \cmidrule(r){3-5} \cmidrule(r){6-8}
\multirow{6}{*}{Learning-based} & \multicolumn{1}{l}{RasterModel~\cite{nuplan}} & 62.93 & 69.66 & \multicolumn{1}{l}{67.54} & 52.4 & 49.47 & \multicolumn{1}{l}{52.16}  \\
 & \multicolumn{1}{l}{UrbanDriver~\cite{urban}} & 82.44 & 63.27 & \multicolumn{1}{l}{61.02} & 76.90 & 51.54 & \multicolumn{1}{l}{49.07} \\
 & \multicolumn{1}{l}{GC-PGP~\cite{pgp}} & 77.33 & 55.99 & \multicolumn{1}{l}{51.39} & 73.78 & 43.22 & \multicolumn{1}{l}{39.63}  \\
 & \multicolumn{1}{l}{PDM-Open~\cite{pdm}} & 84.14 & 52.80 & \multicolumn{1}{l}{57.23} & 79.06 & 33.51 & \multicolumn{1}{l}{35.83} \\
 & \multicolumn{1}{l}{PlanTF~\cite{plantf}} & {87.07} & {86.48} & \multicolumn{1}{l}{{80.59}} & {{83.32}} & {{72.68}} & \multicolumn{1}{l}{{61.70}}  \\
   & \multicolumn{1}{l}{PLUTO~\cite{pluto}} & {\fadedtext{------}} & 89.90 & \multicolumn{1}{l}{78.62} & {\fadedtext{------}} & 70.03 & \multicolumn{1}{l}{59.74}  \\ \cmidrule(r){1-2} \cmidrule(r){3-5} \cmidrule(r){6-8}
 \multirow{5}{*}{Hybrid} 
 & \multicolumn{1}{l}{PDM-Hybrid~\cite{pdm}} & \textbf{82.21} & {90.20} & \multicolumn{1}{l}{{91.56}} & 73.81 & 65.95 & \multicolumn{1}{l}{{75.79}}  \\ 
   & \multicolumn{1}{l}{Diff. Plan~\cite{zheng2025diffusionbased}} & {\fadedtext{------}} & \textbf{94.80} & \multicolumn{1}{l}{\underline{91.75}} & {\fadedtext{------}} & \underline{78.87} & \multicolumn{1}{l}{\textbf{82.00}}  \\
   & \multicolumn{1}{l}{PLUTO~\cite{pluto} w/ refine} & {\fadedtext{------}} & \underline{92.23} & \multicolumn{1}{l}{90.29} & {\fadedtext{------}} & \textbf{80.08} & \multicolumn{1}{l}{76.88}  \\
 & \multicolumn{1}{l}{GameFormer~\cite{gameformer}} & 79.35 & 80.80 & \multicolumn{1}{l}{79.31} & \underline{75.27} & 66.59 & \multicolumn{1}{l}{68.83}  \\
%
   
      &  \multicolumn{1}{l}{\ours  Ours} & \ours   \underline{80.70} & \ours   {91.27} &  \multicolumn{1}{l}{\ours \textbf{92.38}} & \ours   \textbf{76.31} & \ours   {72.08} &   \multicolumn{1}{l}{\ours \underline{77.88}}  \\ 
 \bottomrule
\end{tabular}
\end{center}

\vspace{-1.4em}
\end{table*}

\subsection{Ablation Study}\label{sec:ab}

\subsubsection{Trajectory Entropy Threshold Setting}


\begin{table}[t]
\centering
\caption{Ablation study about Trajectory Entropy threshold. The experiments are conducted on nuPlan planning benchmark. The \colorbox{colorFst}{first}, \colorbox{colorSnd}{second} and \colorbox{colorTrd}{third} results are highlighted.
} 
\label{tab:ab-thd}
\begin{tabular}{ccccc}
\toprule
\multicolumn{2}{c}{Threshold} & \multicolumn{3}{c}{Test14-hard} \\ \cmidrule(r){1-2} \cmidrule(r){3-5}
$T^{0}_{\mathcal{E}}$ & $T^{1}_{\mathcal{E}}$ & OLS & NR-CLS & R-CLS    \\ \midrule
30 & 30 & \rd 75.56	& \fs{71.88}	& \rd 76.04 \\
33 & 33 & \nd 75.75 & \nd 71.86 & \nd 76.67 \\
35 & 35 & \fs{75.77} & \rd 71.85 & \fs {76.94} \\ \midrule
30 & 25 & \fs{76.31} & \fs{72.08} & \rd 75.64 \\
35 & 33 & \rd 74.52 & \nd 71.72 & \nd 77.16 \\
40 & 30 & \nd 75.07 & \rd 71.21 & \fs{77.88} 
\\ \bottomrule
\end{tabular}
\vspace{-1em}
\end{table}

The only manually adjustable parameter in our method is the thresholds $\{T^{i}_{\mathcal{E}}\}_i$ of the Trajectory Entropy Gate.
These thresholds control the strictness of fixing the state of stable agents in the level-k game, with lower values imposing stricter constraints. 
To investigate the impact of $\{T^{i}_{\mathcal{E}}\}_i$ on performance, we conduct planning experiments on the nuPlan Test14-hard benchmark~\cite{plantf}.
The results for six different $\{T^{i}_{\mathcal{E}}\}_i$ settings are presented in \cref{tab:ab-thd}.
Two types of parameter settings are tested.
Setting identical thresholds across different levels results in more fixed agents due to the overall reduction in Trajectory Entropy at deeper levels.
The corresponding results are shown in the first three raws of  \cref{tab:ab-thd}.
Alternatively, decreasing the thresholds with increasing levels aligns better with the reasoning depth, yielding better results, as shown in the last three rows of \cref{tab:ab-thd}. 
Thus, we opt to decrease $\{T^{i}_{\mathcal{E}}\}_i$ with increasing game levels.
The table also reveals that smaller thresholds perform better in the most challenging scenes (NR-CLS planning setting), which fix more agents in the level-k game.
This implies that, in difficult scenes, the model should prioritize fewer but more critical agents in the game for improved planning performance.
Furthermore, we observe that the thresholds on the nuPlan dataset are significantly higher than those on the WOMD. This discrepancy arises because the MTP time horizon in WOMD is 5 seconds, whereas it is 8 seconds in nuPlan. As prediction uncertainty grows significantly with longer time horizons, this validates the effectiveness of our Trajectory Entropy metric in reflecting the driving challenges.

\subsubsection{Trajectory Normalization}
There are various speed limits in different driving scenes.
However, the speed of agents possess less direct relation to driving difficulties.
Thus we utilize trajectory normalization to remove the effect of agent speed on judging the driving difficulties, as shown in \cref{eq:et-f}.
There are different ways to remove the speed effect.
A straightforward approach is to use the expectation of trajectory length at each time step $E^t_M(\mathbf{Y}^t)=\sum\limits_j^M c_j |p_j^t|$ as the normalization factor, which reflect the speed differences between trajectories.
However, the value of $E^t_M(\mathbf{Y}^t)$ increases with the time step, which leads to unbalanced weighting on uncertainty signals from the different time steps.
Alternatively, we can use the trajectory length within a unit time duration $E^{t}_{M}(\mathbf{l})$, representing the instantaneous speed of the agent, as the normalization factor.
From another perspective, as the speed ultimately changes the final lengths of trajectories, we can directly utilize this length expectation $E^t_M(\mathbf{Y})=\sum\limits_j^M c_j |p_j^T|$ as the normalization factor.
To investigate the efficacy of different normalization factors, we conduct experiments on nuPlan in terms of the MTP task. 
We experimentally set proper $\{T^{i}_{\mathcal{E}}\}_i$ for these normalization factors, whose performance is the best among multiple experiment rounds.
The results are reported in \cref{tab:ab-norm}.
It can be seen that an appropriate normalization factor is critical for model performance, as large performance gaps exist in the table.
Overall, the instantaneous speed-related factor $E^{t}_{M}(\mathbf{l})$ obtains the best performance, which can provide balanced weights for uncertainty signals from all time steps.

\begin{table}[!t]
\caption{Ablation study about the normalization factor. The trajectory prediction experiments are conducted on nuPlan benchmark.}
\centering
\resizebox{\linewidth}{!}{
\begin{tabular}{lcccccc}
\toprule
Method   &  $T^{0}_{\mathcal{E}}$ & $T^{1}_{\mathcal{E}}$        & minADE~($\downarrow$)     & minFDE~($\downarrow$)        & Miss Rate~($\downarrow$)   & mAP~($\uparrow$)    \\ \midrule

$E^t_M(\mathbf{Y}^t)$  & 0.03 & 0.025     & {0.4285}          & {0.8252}    & {0.0902}         & \textbf{0.2246}                    \\\midrule

$E^{t}_{M}(\mathbf{l})$  & 40 & 30       & \textbf{0.4180}          & \textbf{0.8063}    & \textbf{0.0898}         & {0.2089}                    \\\midrule

$E^t_M(\mathbf{Y})$   & 0.06 & 0.05      & {0.6602}          & {1.2369}    & {0.0947}         & {0.2037}                    \\

\bottomrule
\end{tabular}%
}
\label{tab:ab-norm}
\vspace{-0.3cm}
\end{table}

\subsubsection{Computational Cost}\label{sec:ab-t-t}

Our Trajectory Entropy is able to avoid unnecessary computation for agents under simple driving environment in the level-k game.
Thus, the computational cost of GameFormer can be reduced through our Trajectory Entropy Gate.
We compare the inference time and GFLOPs of our method with the original GameFormer in the interaction prediction task.
The experiments are conducted on 1 NVIDIA A100 GPU, whose results are summarized in \cref{tab:time}. 
From the table, we can see that the inference time of our method is reduced by $23.84\%$ (38.42 vs. 29.26) compared to the GameFormer, along with a decrease in GFLOPs.
This indicates the proposed Trajectory Entropy enables significant computational efficiency by reducing unnecessary calculations.

\section{Conclusion}

\begin{table}[!t]
\centering
\caption{Computational cost comparison between our method with the original GameFormer.}\label{tab:time}
\begin{tabular}{lll}
\toprule
           & Inference Time (ms) & GFLOPs \\ \midrule
GameFormer  &  38.42   & 1.59   \\
Ours       &  29.26{\imp $_{-23.84\%}$}  & 1.14   \\ \bottomrule
\end{tabular}
\vspace{-1.0em}
\end{table}

This paper introduces Trajectory Entropy to access the game states of agents in recent level-k game framework, for improved precision and efficiency of joint trajectory prediction and planning.
Specifically, we reveal the intrinsic relationship between the MTP uncertainty, agent driving complexity and agent game state in the level-k game framework.
Based on this, we propose to reduce the computation redundancy in the hierarchical game framework, by assigning proper reasoning depth to agents with different driving complexities.
Thus, we extract uncertainty signals from the MTP results in each game level, and utilize the signal-to-noise ratio of the signals to form Trajectory Entropy, as a clue to access the driving complexities.
Through a simple yet effective Trajectory Entropy Gate mechanism, we improve the level-k game framework, reducing the computation overhead and enhancing the overall accuracy at the same time.
Comprehensive experiments are conducted on Waymo and nuPlan benchmarks, in terms of trajectory prediction and open/closed-loop planning, which are key components of intelligent autonomous driving.
The results demonstrate the efficacy of our method, showcasing cutting-edge performance on both benchmarks and enhanced computational efficiency.

{
    \bibliographystyle{IEEEtran}
    \bibliography{IEEEabrv,main}
}

%



%
%
%
%
%

\vfill

\end{document}